\documentclass[10pt,twocolumn,letterpaper]{article}

\usepackage{cvpr}
\usepackage{times}
\usepackage{epsfig}
\usepackage{graphicx}
\usepackage{amsmath}
\usepackage{amssymb}
\usepackage{algorithm}
\usepackage{subfigure}
\usepackage{bm}
\usepackage{times}
\usepackage{wrapfig}
\usepackage{multirow,array}
\usepackage{color}
\usepackage{caption}
\usepackage{epstopdf}

\usepackage[breaklinks=true,bookmarks=false]{hyperref}
 \cvprfinalcopy 

\def\cvprPaperID{1848} 

\ifcvprfinal\fi
\begin{document}

\title{Hybrid-Attention based Decoupled Metric Learning for Zero-Shot Image Retrieval}
\author{Binghui Chen$^{1,}~^{2}$, Weihong Deng$^{1}$\thanks{Corresponding author}\\
$^{1}$Beijing University of Posts and Telecommunications\\
$^{2}$AI Labs, Didi Chuxing, Beijing 100193, China\\
{\tt\small chenbinghui@bupt.edu.cn, whdeng@bupt.edu.cn}
}
\maketitle
\thispagestyle{empty}
\vspace{-2em}
\begin{abstract}
   In zero-shot image retrieval (ZSIR) task, embedding learning becomes more attractive, however, many methods follow the traditional metric learning idea and omit the problems behind zero-shot settings. In this paper, we first emphasize the importance of learning visual discriminative metric and preventing the partial/selective learning behavior of learner in ZSIR, and then propose the Decoupled Metric Learning (DeML) framework to achieve these individually. Instead of coarsely optimizing an unified metric, we decouple it into multiple attention-specific parts so as to recurrently induce the discrimination and explicitly enhance the generalization. And they are mainly achieved by our object-attention module based on random walk graph propagation and the channel-attention module based on the adversary constraint, respectively. We demonstrate the necessity of addressing the vital problems in ZSIR on the popular benchmarks, outperforming the state-of-the-art methods by a significant margin. \textcolor[rgb]{1,0,0}{Code is available at http://www.bhchen.cn}
\end{abstract}
\vspace{-2em}
\section{Introduction}\label{introduction}
\vspace{-0.2em}
In \emph{zero-shot image retrieval} (ZSIR), \textbf{the model is required to learn embedding from the \emph{seen} classes and then to be capable of utilizing the learned knowledge to distinguish the \emph{unseen} classes without any attributes or semantic information}.
Most existing works adopt deep embedding (metric) learning and dedicate to exploring powerful loss functions, such as pair-based methods \cite{sun2014deep,Yuan_2017_ICCV,Schroff2015FaceNet,oh2016deep,Sohn2016npair,chen2018almn,yuan2019SNR}, and hard-example mining strategies \cite{kumar2017smart,Wu_2017_ICCV,Schroff2015FaceNet,Sohn2016npair}, \emph{etc}. These ideas can be summarised as: \emph{learn a good matric on the \emph{seen} categories so as to boost the performance on the \emph{unseen} ones}. However, in fact, they neglect the existence and significance of some vital problems to ZSIR, thus are easily stuck in the particular data distribution and knowledge zone of the \emph{seen} training set, some helpful knowledge for \emph{unseen} classes may have been left out with a high probability, as a result, their performances are almost on par with each other and unsatisfactory.
\begin{figure}[t]
   \begin{center}
  \vspace{-1.5em}
    \includegraphics[width=0.9\linewidth]{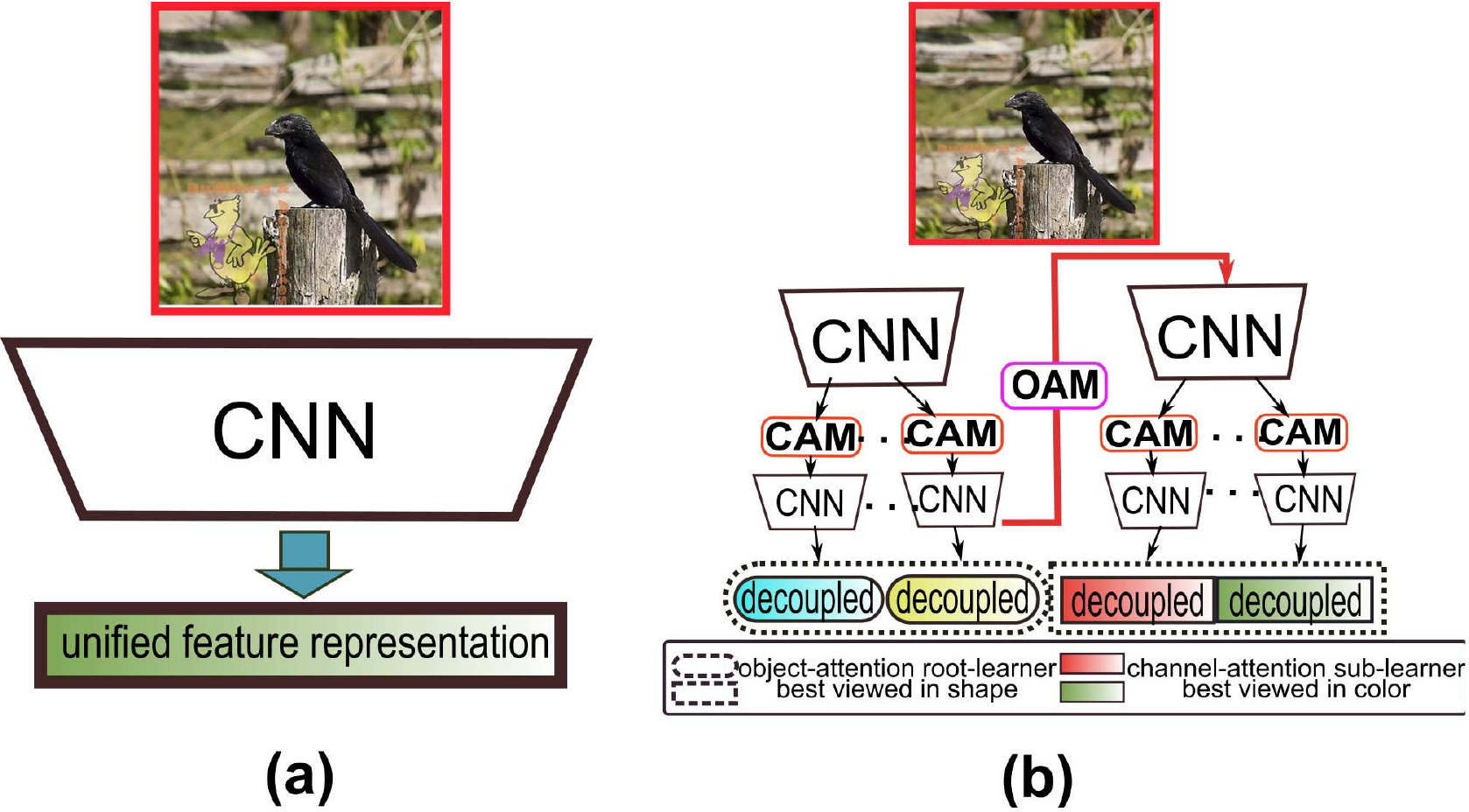}
  \end{center}
  \vspace{-1.7em}
  \captionsetup{font={scriptsize}}
  \caption{Differences between \textbf{(a)} unified metric learning and \textbf{(b)} DeML. Our DeML decouples the unified representations into multiple attention-specific learners so as to encourage the discrimination and generalization of the holistic metric.}\label{fig_decoupled}
  \vspace{-1.7em}
\end{figure}

Specifically, in ZSIR, the ideas above are actually unreasonable as the discrimination and generalization abilities of the learned metric are highly influenced by the following two aspects: \textbf{(1)} the undiscriminating visual input, which contains much noisy information (it is hard to know where should be focused on), resulting in poor descriptors and inducing difficulty in distinguishing the \emph{unseen} classes. In another word, undiscriminating visual input deteriorates the discrimination of features. \textbf{(2)} partial/selective learning behavior of deep models\cite{chen2019energy}, specifically, for a functional learner parameterized by CNN, when given an arbitrary input, it will selectively learn partial attributes knowledge that are the easiest ones to decrease the current training empirical risk over the \emph{seen} categories, instead of learning all-sided details and information, thus yielding over-fitting on \emph{seen} categories and generalizing worse to the \emph{unseen} ones. For example on birds, if the head knowledge is enough to distinguish the training \emph{seen} classes, the deep model will only focus on the head and ignore the other body parts like back, foot and wing, while when testing on the \emph{unseen} classes (which have similar head features but different wing features), the trained model will easily fail to distinguish them\footnote{Since in zero-shot settings, the unseen classes have no intersection with the seen classes, this partial learning behavior is common.}. This partial learning behavior sometimes is one reason for neural redundancy, since all the neurons only focus on the certain easy knowledge, causing that the performance doesn't increase much with the feature dimension. Moreover, when optimizing an unified metric as in the aforementioned methods (as in Fig.\ref{fig_decoupled}.(a)), these two issues are commonly mixed with each other and jointly impede the learning of representation, and without explicit and helpful guidance, the poor discrimination and generalization characteristics of features are greatly exacerbated regardless of the efficacy of objective function and hard-sample mining strategy. And most ZSIR works ignore the importance of learning both discriminative and robust discriminators. To this end, proposing knowledge-preserving and discriminative metric learning framework remains important.

In this paper, inspired by the idea of ``divide and rule'', we propose the \emph{\textbf{Decoupled Metric Learning}} (DeML), a novel and generally applicable framework, to alleviate the above problems by decoupling the embedding representation into multiple attention-specific learners so as to encourage its discrimination and generalization in an explicit manner, as shown in Fig.\ref{fig_decoupled}.(b). Moreover, equipping each learners with the particular attention tasks will further alleviate the neural redundancy and give chances of performance improvement.
Our DeML is a model-agnostic learning framework, and different from the existing metric learning works, it has no need to mine hard samples or construct elaborate sample-pairs. The main contributions are as follows:

\begin{small}
$\bullet$ We provide insights of two vital problems to ZSIR that \emph{undiscriminating visual input} and \emph{partial/selective learning behavior of learner} will impede the discrimination and generalization of the metric, \emph{resp}. And propose DeML, a hybrid-attention based decoupling method, to address these challenges individually.

$\bullet$ DeML decouples the commonly used unified metric into object-attention root-learners and channel-attention sub-learners, aiming at recurrently inducing the discrimination and explicitly promoting the diversity together with the generalization, \emph{resp}.

$\bullet$ The decoupled learners are supported by the corresponding \emph{random walk inferred object-attention modules (OAM)}\footnote{Some works call it spatial-attention.} and the \emph{adversarial channel-attention modules (CAM)}. They can be easily performed as the drop-in parts and are model--agnostic.

$\bullet$ We demonstrate the necessity to learn discriminative and robust metric via our decoupling idea. Extensive experiments have been performed on several challenging datasets for ZSIR, including CUB \cite{Wah2011The}, CARS \cite{Krause20133D}, Stanford Online Products \cite{oh2016deep} and In-Shop \cite{liu2016deepfashion}. Our DeML achieves state-of-the-art performances and surpasses other methods by a significant margin.
\end{small}
\vspace{-0.8em}
\section{Related Work}
\vspace{-0.3em}
\begin{figure*}[t]
  \centering
  \includegraphics[width=0.7\linewidth]{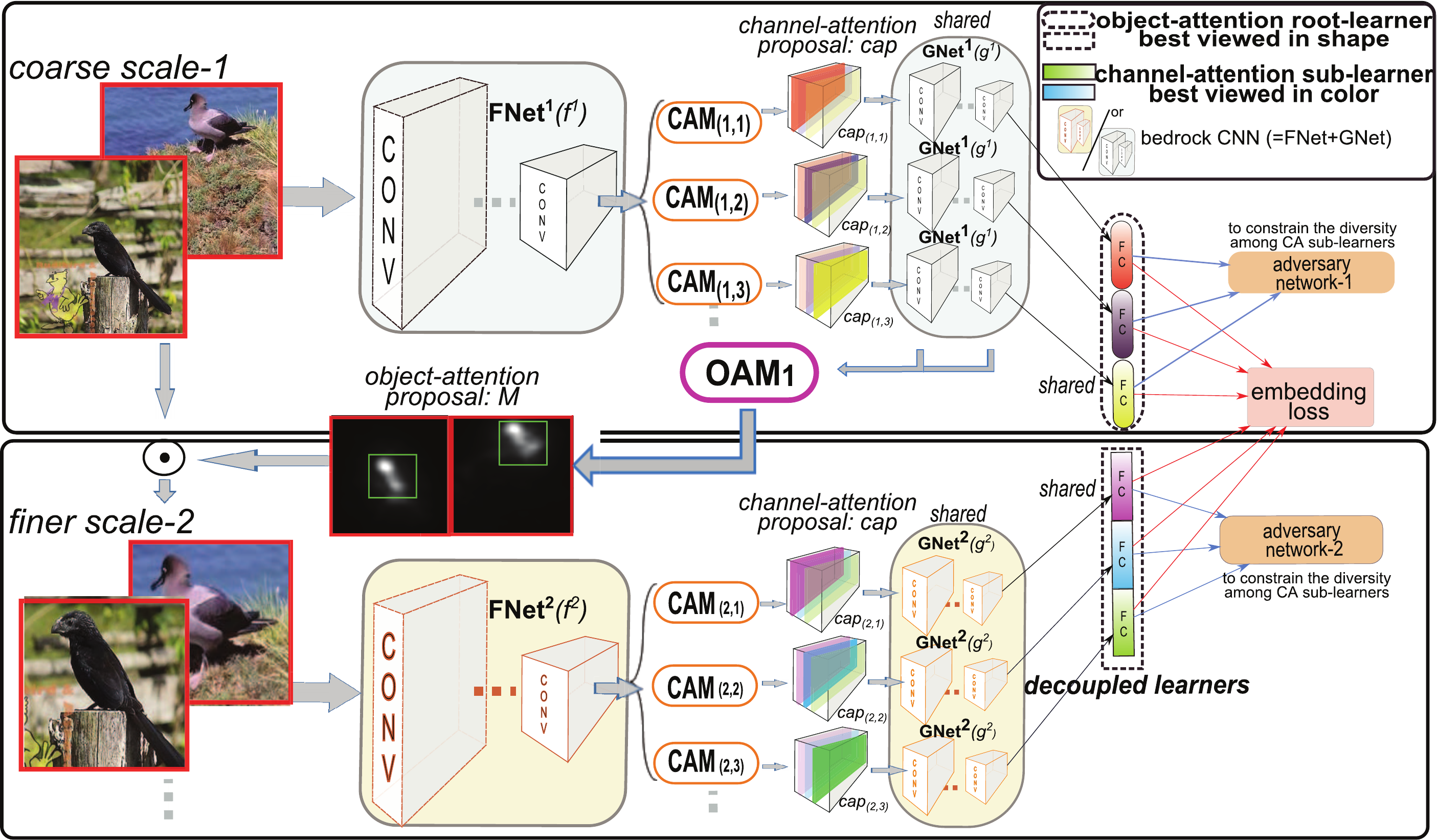}\\
  \vspace{-0.6em}\captionsetup{font={footnotesize}}
  \caption{The framework of our DeML. $\bigodot$ indicates the joint operation of cropping and zooming. The FC layer is first decoupled into two object-attention root-learners(dashed rectangle and ellipse) for coarse and finer scale, \emph{resp}. Then, each root-learner is further decoupled into three channel-attention sub-learners (best viewed in colors). Each (root or sub) learner is supported by the corresponding attention module (OAM or CAM). The parameters to be shared across streams have been annotated by 'shared'. ($FNet$+$GNet$) is the whole GooglenetV1, and we use $FNet^{i},GNet^{i}$ to indicate the nets at scale $i$. Details of the adversary net is in Fig.\ref{fig_graph_adversary}.(c)}\label{fig_flowchart}
  \vspace{-1.8em}
\end{figure*}
\textbf{Zero-shot learning}: ZSL has been widely studied in many tasks, such as image classification \cite{li2018discriminative,changpinyo2016synthesized,fu2015transductive,zhang2015zero}, hashing\cite{shen2018zero}, video recognition \cite{dalton2013zero} \emph{etc}. However, different from these ZSL tasks that are capable of exploiting the extra auxiliary supervision signals of \emph{unseen} classes (e.g. semantic word embedding of class name and explicit attribute information), \emph{\textbf{our DeML concentrates on a more general version of zero-shot image retrieval (ZSIR), where there are only similarity labels $s_{pq}\in\{0,1\}$ available}}. Therefore, how to capture the discriminative and all-sided information only from the input image is the mainstay of this task.

\textbf{Deep embedding/metric learning}: With only $s_{pq}$ available, many works seek to exploring deep metric learning for ZSIR task. For example, sampling-Matters \cite{Wu_2017_ICCV} proposes distance weighted sampling strategy. Proxy-NCA \cite{Movshovitz-Attias_2017_ICCV} explains why popular classification loss works from a proxy-agent view, and its implementation is very similar with Softmax. ALMN \cite{chen2018almn} proposes to optimize an adaptive large margin objective via the generated virtual points instead of mining hard-samples. However, all the above methods are to tackle with the unified metric by designing losses and exploring sample-mining strategies, thus suffer from the aforementioned issues easily. Additionally, HDC \cite{Yuan_2017_ICCV} employs the cascaded models and selects hard-samples from different levels and models. BIER loss \cite{Opitz_2017_ICCV,opitz2018deep} adopts the online gradients boosting methods. Although these two methods try to improve the performances by resorting to the ensemble idea, they can only learn from the undiscriminating input and also suffer from the partial learning behavior, as a result, the discrimination and generalization of holistic features are still limited.

\textbf{Attention mechanism}: Attention serves as a tool to bias the allocation of available resources towards the most informative parts of an input. Many are implemented in combination with a gating mechanism (e.g. softmax or sigmoid) and are widely applied in many tasks, e.g. image captioning \cite{chen2017sca,xu2015show}, lip reading \cite{chung2016lip}, image classification \cite{hu2017squeeze,Wang_2017_CVPR,fu2017look}. Schwartz et at. \cite{schwartz2017high} adopt high order attention modules for VQA. Fu et at. \cite{fu2017look} propose to learn an attention network so as to produce the attention proposals, while, it is optimized in a two-stage manner, i.e. iteratively train the embedding model and attention model. We emphasize that our DeML is a hybrid-attention system. First, it adopts a parameter-free \emph{object-attention module}, which is performed by random walk graph propagation and can be directly plugged into any bedrock CNNs without training. Second, different from \cite{hu2017squeeze,chen2017sca} where the channel-attention operations are cascaded, i.e. only the most informative signal will be captured, our DeML is equipped with a series of \emph{channel-attention modules} in parallelism, which are optimized by the adversary module and insist to learn various combinations of attributes simultaneously and to capture knowledge as much as possible.
\vspace{-0.5em}
\section{Decoupled Metric Learning}
\vspace{-0.3em}
As discussed in Sec.\ref{introduction}, in \emph{zero-shot} setting, simply optimizing an unified representation like in most existing works wouldn't guide the learning of discriminative and robust metric as a joint result of (1) undiscriminating visual input and (2) the partial/selective learning behavior\footnote{\footnotesize{Notably, these shortcomings also reside in conventional full-shot tasks, while they heavily exacerbate the performances in zero-shot settings.}}.

\textbf{\emph{Naturally motivated by ``divide and rule'', the intuitive solution is to decouple the single representation into a series of artificially defined learners so as to explicitly encourage different learners to attend to different latent features and improve the holistic feature's discrimination and diversity}}. Moreover, many research works point out that attention mechanism is a crucial component connecting data embeddings with decision making modules, and diverse attentional cues allows the learning of various knowledge. We follow this inspiring idea and recast the above solution as \emph{hybrid-attention} learning by decoupling the original unified metric into several attention-specific parts, each part is followed by the corresponding fc learner, i.e. \emph{object-attention} (OA) root-learner and \emph{channel-attention} (CA) sub-learner. The overall framework of DeML is illustrated in Fig.\ref{fig_flowchart} (for clarity, we depict the network with only two OA root-learners and three CA sub-learners under each root-learner as an example).

Specifically, for the $i$-$th$ scale, the embedding part consists of three different components: (1) the bedrock CNN, i.e. Googlenet-V1 \cite{Szegedy2014Going} (we decompose it into $\frac{(Conv1-Pool3)}{FNet}\frac{(Incep4-Pool5)}{GNet}$), (2) the \emph{object-attention module} (OAM) is to incrementally pinpoint the informative object-based region proposals for discriminative feature learning, and (3) the \emph{channel-attention module} (CAM) is to concurrently produce diverse attribute-based channel-attention proposal $cap_{(i,j)}$. When given an image $x_{n}$, we will directly use it for the first scale input and employ its cropped\&zoomed version for the next scale input. At the $i$-$th$ scale, the input will first be encoded by \emph{FNet$^{i}$} and then fed into multiple CAMs. The produced \emph{channel-attention} proposals $cap_{(i,j)}$ will then be encoded by the corresponding subsequent \emph{GNet$^{i}$} and CA sub-learners. All the CA sub-learners at the same scale constitute one OA root-learner. We now define $OA_{i},CA_{(i,j)}$ as the corresponding learners and $OA_{i}(\cdot),CA_{(i,j)}(\cdot)$ as the corresponding encoding functions, then give the holistic feature representation $\varphi[x_{n}]$ as:

\vspace{-1.5em}
\begin{small}
\begin{align}\label{eq1}
&\varphi[x_{n}]=\{OA_{1}[x_{n}^{1}];~OA_{2}[x_{n}^{2}];~\cdots\}\\
w&here~OA_{i}[\cdot]=\{\cdots;~CA_{(i,j)}[\cdot];~\cdots\}\nonumber
\end{align}
\end{small}
\vspace{-1.5em}

where the input $x_{n}\in\mathcal{X}_{seen}\equiv\{x_{n}\}_{n=1}^{N}$, $x_{n}^{i}$ is the $i$-th scale image which is cropped and zoomed from $x_{n}^{i-1}$ by $OAM_{i-1}$, and $x_{n}^{1}=x_{n}$ is the original scale image. Notably, we use square brackets ``[]'' to show the conditional relation, e.g. $CA_{(i,j)}[x_{n}^{i}]$ indicates the CA learner $CA_{(i,j)}$ is conditional on input $x_{n}^{i}$, and this relation can also be formulated as $CA_{(i,j)}[x_{n}^{i}]=CA_{(i,j)}(g^{i}(CAM_{(i,j)}(f^{i}(x_{n}^{i}))))$ (shows the encoding process from image to feature space), where $f^{i}$ and $g^{i}$ indicate the encoding functions of \emph{FNet$^{i}$} and \emph{GNet$^{i}$}, \emph{resp}, $CAM_{(i,j)}$ denotes the corresponding \emph{channel-attention module} (CAM), $i\in [1\cdots I], j\in [1\cdots J]$ (in Fig.\ref{fig_flowchart}, $I$=$2,J$=$3$). The total size of $\varphi{[x_{n}]}$ is $d$-dim, thus for $I$ OA learners and $I\times J$ CA learners, $OA_{i}(\cdot)\in \mathbb{R}^{\frac{d}{I}}$ and $CA_{(i,j)}(\cdot)\in \mathbb{R}^{\frac{d}{I\times J}}$.

Below, we will introduce our OAM and CAM in Sec.\ref{OAM} and Sec.\ref{CAM} \emph{resp}, and then show the objective metric loss together with the regularization term in Sec.\ref{overall}.
\vspace{-0.5em}
\subsection{Random Walk Inferred OAM}\label{OAM}
\vspace{-0.3em}
\begin{figure*}[t]
  \centering
  \includegraphics[width=0.7\linewidth]{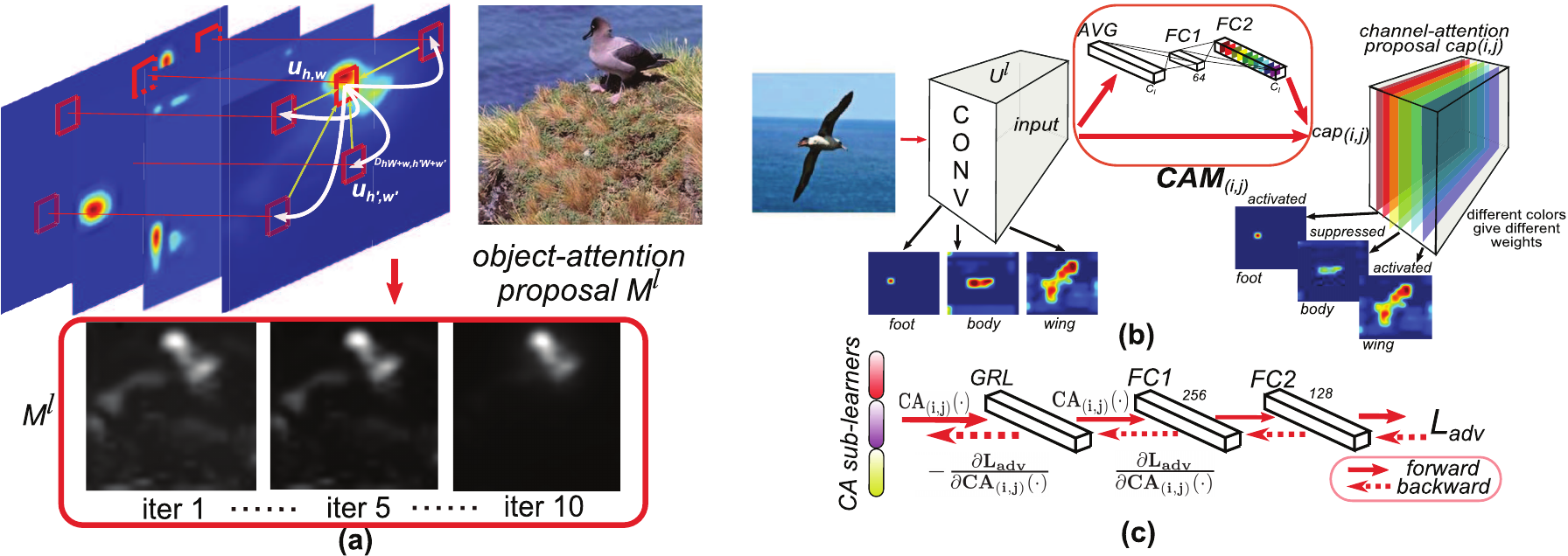}\\
  \vspace{-1em}\captionsetup{font={footnotesize}}
  \caption{(a) The generation of object-attention proposal $M^{l}$ via random walk graph propagation. (b) The generation of channel-attention proposal $cap_{(i,j)}$ via CAM, in this case, 'foot' and 'wing' are activated while 'body' is suppressed, and if there are diverse CAMs , different cases will turn up, (c) The adversary network, which is placed after the CA sub-learners.}\label{fig_graph_adversary}
  \vspace{-1.7em}
\end{figure*}
As we discussed above, in order to learn discriminative OA learner, we should first provide the discriminative visual inputs. However, unlike the detection methods that have precise bounding-box annotations about object locations(e.g. RPN \cite{girshick2015fast,ren2015faster}), there are no auxiliary annotations for training in ZSIR. We thus present the random walk inferred \emph{object-attention module}(OAM), a parameter-free and spatial attention method, by performing graph propagation over the receptive fields of deep responses in Conv maps, so as to produce the object-attention proposals.

As illustrated in Fig.\ref{fig_flowchart}, our OAM takes as input many convolutional response maps ($U^{l_{k}}\in \mathbb{R}^{C_{l_{k}}\times{H_{l_{k}}}\times{W_{l_{k}}}}$, $k\in{[1,2,\cdots]}$, $C_{l_{k}}, H_{l_{k}}$ and $W_{l_{k}}$ denote the number of channels, height and width of the $l_{k}$-th layer, respectively), and then gives the final object-attention proposal. For clarity, we give a toy example and omit the subscript $k$ for written convenience (if not specified), each spatial location $(h,w)$ on $U^{l}$ has a feature vector $u_{h,w}^{l}\in{\mathbb{R}^{C_{l}}}$. To produce the spatial object-attention proposal $M^{l}\in{\mathbb{R}^{H_{l}\times{W_{l}}}}$, a fully connected directed graph $G^{l}$ is first constructed by connecting every location on $U^{l}$ with the weight matrix $D^{l}\in{\mathbb{R}^{H_{l}W_{l}\times{H_{l}W_{l}}}}$, where $D_{hW_{l}+w,h^{'}W_{l}+w^{'}}^{l}$ indicates the weight of the edge from node $(h,w)$ to $(h^{'},w^{'})$ and is defined as:

\vspace{-1.3em}
\begin{small}
\begin{align}\label{eq2}
  D_{hW_{l}+w,h^{'}W_{l}+w^{'}}^{l}&\triangleq\|u_{h,w}^{l}-u_{h^{'},w^{'}}^{l}\|
\end{align}
\end{small}
\vspace{-1.2em}

Before performing propagation, the weights of the outbound edges of each node are normalized to $1$, i.e. $D_{a,b}^{l}\leftarrow\frac{D_{a,b}^{l}}{\sum_{b^{'}=1}^{H_{l}W_{l}}D_{a,b^{'}}^{l}}$. Then, a random walk graph propagation algorithm \cite{lovasz1993random} is employed to produce the object-attention proposal $M^{l}$. As shown in Fig.\ref{fig_graph_adversary}.(a), the node $(h,w)$ receives and diffuses mass via the inbound and outbound directed edges respectively. Then this random walk propagation iteratively collects mass at the nodes that have high dissimilarity with their surroundings (i.e. it will highlight the saliency foreground object region), since the edge weight (i.e. transition probability) is proportional to the dissimilarity between two nodes as defined by Eq.\ref{eq2}, transitions into such subgraphs is likely, and unlikely if nodes have similar deep features $u^{l}$. By the globally objectness flow, it can not only collects object evidence but also depresses noise regions. For the convenience of random walk calculation, the proposal map $M^{l}$ is first reshaped to a $H_{l}W_{l}$-dim vector, and then it is updated by the iterative multiplication with the weight matrix $D^{l}$, i.e. $M\leftarrow{D\times{M}}$.

Moreover, as a common knowledge that the response maps at different layers are of different information, e.g. the bottom layers have better predictive power for shape while the top layers are robust to noisy regions \cite{long2015fully,ronneberger2015u}, our OAM takes as input multiple Conv. layers $U^{l_{k}},k\in{[1,2,\cdots]}$ so as to avoid the influence of noise surroundings (e.g. trees around birds), and each object-attention proposal $M^{l_{k}}$ can be obtained by performing random walk graph propagation over $U^{l_{k}}$ as follows:
\vspace{-1.2em}
\begin{equation}\label{eq3}
  \small{M^{l_{k}}\leftarrow(\prod_{t=1}^{T}D^{l_{k}})M^{l_{k}}}
\end{equation}
\vspace{-1em}

$M^{l_{k}}$ is first initialized with value $\small{\frac{1}{H_{l_{k}}W_{l_{k}}}}$, and the weight matrix $D^{l_{k}}$ is conditional on the response map $U^{l_{k}}$, then the iterative update of $M^{l_{k}}$ like $M\leftarrow{D\times{M}}$ can be rewritten as the above equation and each $M^{l_{k}}$ can reach its stable state with $T=10$. After Eq.\ref{eq3}, all $M^{l_{k}}$ are reshaped to $2D$ maps and resized to the same size as the input image, then they are averaged to a single map, i.e. the final object-attention proposal. In experiments, we find using \{$k\in[1,2], and~ l_{1}=Incep4e, l_{2}=Incep5b$\} is enough, As in Fig.\ref{fig_flowchart}, with the help of this proposal, the coarse scale image is processed into the finer scale by operation $\bigodot$, and then the finer scale input will benefit the discriminative learning of the corresponding OA learner.

\textbf{Remark}: The procedure in Eq.\ref{eq3} can be viewed as a variant of the eigenvector centrality measure \cite{newman2008mathematics}, which outputs a proposal map to indicate the mass of each location on the deep response maps. It can also be regarded as a Markov chain that can reach unique stable state since the chain is ergodic, a property which emerges from the fact that the graph $G$ is by construction strongly connected. \cite{harel2007graph,zhu2017soft}.
Moreover, the random-walk based informative-region search is computationally lightweight and parameter-free, thus it can be easily employed in any bedrock CNNs and has no need to be additionally optimized, i.e. when given the well trained convolutional maps $U$, it can directly infer the important object region in an unsupervised manner. Finally, recurrently executing OAMs will incrementally improve the discrimination of the corresponding OA learners.
\vspace{-0.3em}
\subsection{Adversarial CAM}\label{CAM}
\vspace{-0.2em}
Although the OA learners can produce discriminative representations, each single OA learner will still be likely to suffer from the partial/selective learning behavior, i.e. only focuses on certain attributes (e.g. heads of birds, bellows of car, \emph{etc}.) which are the easiest ones to reduce the current training risk. Therefore, in order to improve the diversity of each OA root-learner, we further decouple it into $J$ CA sub-learners which insist to capture various attributes information coming from diverse channel-attention proposals $cap_{(i,j)}$ as in Fig.\ref{fig_flowchart}, and each $cap_{(i,j)}$ is produced by a particular \emph{channel-attention mudule} (CAM$_{(i,j)}$). For clarity, we take one CAM for example, as shown in Fig.\ref{fig_graph_adversary}.(b), it takes as input $U^{l}\in{\mathbb{R}^{C_{l}\times{H_{l}}\times{W_{l}}}}$ of the $l$-th layer (omit subscript $k$ for convenience) and outputs a channel-weighted counterpart of $U^{l}$, i.e. $cap_{(i,j)}$. We formulate it as:
\begin{small}
\begin{align}
  cap_{(i,j)}=CAM_{(i,j)}(U^{l})=U^{l}\boxdot \sigma[W_{(i,j)}^{2}\delta(W_{(i,j)}^{1}\Psi(U^{l}))]\nonumber
\end{align}
\end{small}
where $U^{l}$ is first aggregated to a channel descriptor through spatial average pooling $\Psi$, then is passed through two fully-connected layers (parameterized by $W_{(i,j)}^{1}\in{\mathbb{R}^{64\times{C_{l}}}}, W_{(i,j)}^{2}\in{\mathbb{R}^{C_{l}\times64}}$) that intend to capture the interactions among channels and are followed by $ReLU$ ($\delta$) and $Sigmoid$ ($\sigma$) respectively. Finally, $U^{l}$ is reweighted by the channel-wise multiplication $\boxdot$ between $2D$ map $U^{l}_{c}$ and scalar $\sigma[W_{2}\delta(W_{1}\Psi(U^{l}))]_{c}$, where $c$ is the channel index.

Since the filters in certain convolutional layers serve as the local attribute detectors to some extend, the resulting deep response maps will reserve the attribute information in channels. In another word, different attributes reside in different channels. Therefore, the reweighting operation in our CAM can be regarded as an attributes selector(e.g. in Fig.\ref{fig_graph_adversary}.(b) it selects 'foot' and 'wing' yet ignores 'body'), and the concerned cues(within $cap_{(i,j)}$) will be passed to the corresponding CA sub-learner. If there are diverse selectors, i.e. multiple CAMs, capable of capturing different combinations of attributes information in parallel, the corresponding CA learners will focus on different attributes knowledge, and the richness and diversity of each OA root-learner will be significantly improved.

While, to explicitly encourage the diversity of CAMs, directly adding diversity constraints after CAMs might be detrimental, since in fact it is hard to know which channel contains what attribute signal and which kind of channel combination is beneficial to the embedding learning. To this end, as shown in Fig.\ref{fig_flowchart}, all $cap_{(i,j)}$ are first fed into the corresponding subsequent \textbf{\emph{shared}} embedding layers, and the diversity of CAMs are naturally transferred to that of the CA learners. And in order to explicitly encourage the diversity of CAMs, we can instead impose diversity constraints on the CA learners. Then, an adversarial optimization strategy is introduced as the diversity constraint by playing a $\max\min$ game, i.e. the adversary network tries to minimize the discrepancies among CA learners while CAMs try to maximize these discrepancies as follows:
\vspace{-0.5em}
\begin{small}
\begin{align}
  \max_{f^{i},g^{i},CAMs}\min_{\digamma^{i}} L_{adv}(x_{n}^{i};\digamma^{i};f^{i};g^{i} ;CAMs)=\nonumber
\end{align}
\vspace{-1.6em}
\begin{align}\label{eq5}
  \lambda_{0}\sum_{j,j^{'}}^{J}\|\digamma^{i}[CA_{(i,j)}[x_{n}^{i}]]-\digamma^{i}[CA_{(i,j^{'})}[x_{n}^{i}]]\|^{2}_{2}
\end{align}
\end{small}
\vspace{-1em}

where $f^{i},~g^{i}$ and $\digamma^{i}$ indicate the mapping functions of $FNet^{i}$, $GNet^{i}$ and the adversary network respectively, $\lambda_{0}$ is the loss weight. The above equation is a discrepancy measurement over all $J$ CA sub-learners that belong to the $i$-th OA root-learner. Moreover, to simplify the two-stage optimization in this $\max\min$ game, a \emph{gradients reverse layer} (GRL) \cite{ganin2016domain} is introduced as shown in Fig.\ref{fig_graph_adversary}.(c), so as to make Eq.\ref{eq5} be equivalent to $\min_{\digamma^{i}}L_{adv}$ and $\min_{(f^{i},g^{i},CAMs)} (-L_{adv})$.

\textbf{Remark}: Since at each scale $i$ the params of $GNet^{i}$ ($g^{i}$) in different channel-attention branches are shared as shown in Fig.\ref{fig_flowchart}, when optimizing $\min_{(f^{i},g^{i},CAMs)} (-L_{adv})$, the only differentiating parts are CAMs, in other words, only diverse CAMs will make $(-L_{adv})$ small. Moreover, the reason of using adversary objective instead of directly constrain CA learners to be different with each other is because the definition of difference between learners is likely to be hard to be artificially made than the similarity, we thus play the above adversarial game so as to allow the automatic learning of the optimal discrepancies between CAMs.
\vspace{-0.5em}
\subsection{Objective Loss and Regularization Term}\label{overall}
Employing multiple OAMs and CAMs, the holistic feature representation $\varphi[x_{n}]$ can be decoupled into a series of attention-specific parts. Then, the binomial deviance loss \cite{yi2014deep} is utilized and the empirical metric loss over all CA sub-learners can be formulated as:
\vspace{-0.5em}
\begin{equation}\label{eq6}
  \small{L_{metric}=\frac{1}{IJ}\sum_{i=1}^{I}\sum_{j=1}^{J}\sum_{p,~q}\frac{1}{w_{pq}}\log{(1+e^{-(2s_{pq}-1)\alpha(D^{(i,j)}_{pq}-\beta)\gamma_{pq}})}}
\vspace{-0.5em}
\end{equation}
where $\alpha=2, \beta=0.5$ are the scaling and translation parameters \emph{resp}. $\gamma_{pq}$ is the penalty coefficient and is set to $1$ if $s_{pq}=1$(i.e. images $x_{p},x_{q}$ are from the same class), otherwise $\gamma_{pq}=35$. $w_{pq}$ is the number of positive pairs(negative pairs) if $s_{pq}=1$($s_{pq}=0$). $D_{pq}^{(i,j)}=\frac{<CA_{(i,j)}[x_{p}^{i}],CA_{(i,j)}[x_{q}^{i}]>}{\|CA_{(i,j)}[x_{p}^{i}]\|\|CA_{(i,j)}[x_{q}^{i}]\|}$ measures the distance between inputs $x_{p}^{i}$ and $x_{q}^{i}$ at learner $CA_{(i,j)}$. Notably, if $J=1$, the OA root-learner has no need to be decoupled, thus $CA_{(i,j)}=OA_{i}$. Additionally, the learners are trained on the relatively small datasets (compared to ImageNet \cite{russakovsky2015imagenet}) from scratch and thus are easier to be over-fitting, following the commonly used weight decay strategy which regularizes the value of parameters not to be too large so as to reduce over-fitting, we present a regularization term for optimizing each learner, called \emph{activation decay}, to constraint the representation value as follows:

\vspace{-1.2em}
\begin{footnotesize}
\begin{align}\label{eq7}
  &L_{act}=\frac{\lambda_{1}}{2IJN}\sum_{i=1}^{I}\sum_{j=1}^{J}\sum_{n=1}^{N}\|CA_{(i,j)}[x_{n}^{i}]\|^{2}_{2}+L_{ntri}\\
  \vspace{-0.6em}
  &where~L_{ntri}=\lambda_{2}\sum_{i=1}^{I}\sum_{j=1}^{J}tr[(\omega_{ij}\omega_{ij}^{T}-I)\otimes(\omega_{ij}\omega_{ij}^{T}-I)]\nonumber
\end{align}
\end{footnotesize}
\vspace{-1em}

\noindent where $L_{ntri}$ is used to avoid the trivial solution(i.e. all parameters are optimized to $0$) when minimizing $L_{act}$, $\lambda_{1},\lambda_{2}$ are trade-off hyper-parameters. $\omega_{ij}\in{\mathbb{R}^{\frac{d}{IJ}\times{d_{1}}}}$ are the parameters in learner $CA_{(i,j)}$, $d_{1}$ indicates its input channel number. $\otimes$ refers to the Hadamard-Product. Finally, DeML can be trained by jointly optimizing Eq.\ref{eq5}, \ref{eq6} \& \ref{eq7}.

\begin{figure*}[t]
\centering
  \includegraphics[width=0.9\linewidth]{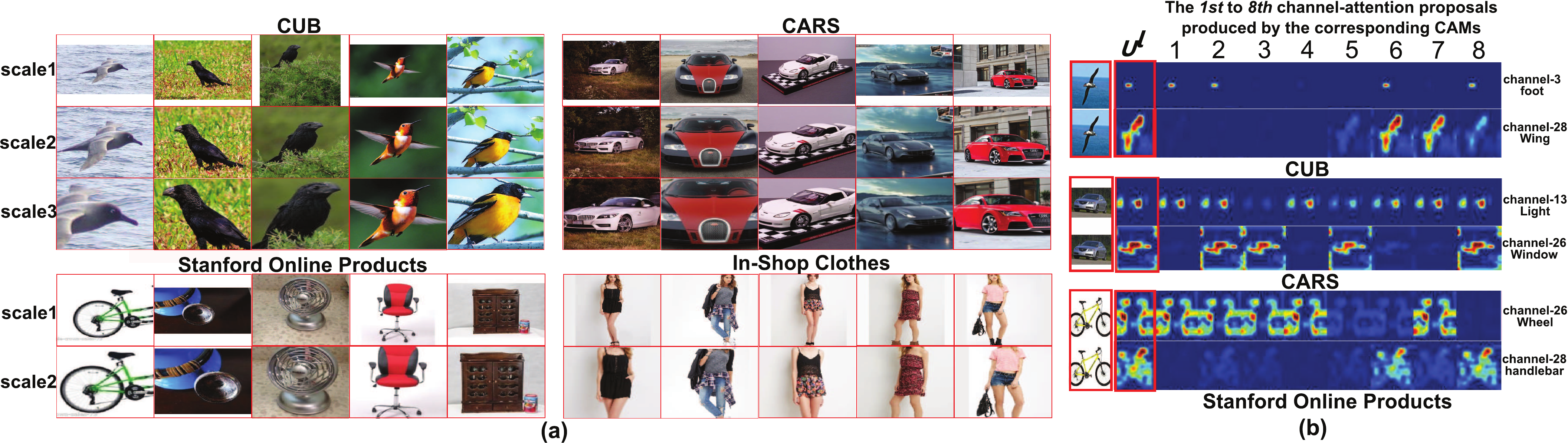}\\
  \vspace{-0.7em}\captionsetup{font={footnotesize}}
  \caption{(a) Object-attention regions at different scales inferred by OAMs. (b) Channel-attention proposals at certain channels output by CAMs. }
\label{fig4}
  \vspace{-2.1em}
\end{figure*}
\vspace{-0.5em}
\subsection{Discussion}
\vspace{-0.5em}
\textbf{Why decoupling?} In ZSIR, the traditional idea of devising loss functions for optimizing an unified metric omits the issues of (1) undiscriminating visual input regions and (2) the partial/selective learning behavior of learner, and thus exhibits limitations in discriminative and knowledge-preserving feature learning that is crucial for ZSIR. However, in our DeML, decoupling this unified metric into multiple separate learners gives us chances and flexibilities to mitigate the above problems individually, i.e. recurrently improve the discrimination via OA root-learners and explicitly encourage the diversity and richness via CA sub-learners, and as a result, the redundancy in features is reduced and the representation ability is encouraged as well. Additionally, in Sec.\ref{experiments}, we show that our decoupling idea outperforms the existing methods in ZSIR by a significant margin even without hard-samples mining.

\textbf{Relation to other ensemble methods}: 
In deep metric learning community, there are some other ensemble works related to us. Yuan et al \cite{Yuan_2017_ICCV} employ multiple layers at different depths for hard-aware samples mining and then cascade the learned embeddings together. Opitz et at \cite{Opitz_2017_ICCV,opitz2018deep} adopt the online gradients boosting and optimize different learners with the reweighted data. Kim et at \cite{Kim_2018_ECCV} try to increase the feature diversity via contrastive loss but ignore the importance of learning discriminative metric in ZSIR task. In summary, although these methods aim at learning ensemble metric, they easily suffer from the undiscriminating inputs, and the partial learning behaviors are not effectively constrained as well. We emphasize that our DeML recasts the ensemble problem as a hybrid-attention model that has a clear object of addressing the aforementioned issues individually, and decouples the unified metric into multiple attention-specific parts, so as to explicitly intensify both the discrimination and diversity within each part of features, thus capturing rich knowledge and being capable of generalizing to unseen categories.

\textbf{Relation to other attention mechanism}: Many other attention works, such as spatial-attention \cite{xu2015show,fu2015transductive}, channel-attention \cite{chen2017sca,hu2017squeeze} and semantic-attention \cite{you2016image,schwartz2017high}, employ the task-specific attention modules to mine the most informative signals. While, we emphasize that our DeML is a hybrid-attention system, which dedicates to capturing not only the informative object-attention regions via OAMs but also the diversity of attributes via CAMs. Moreover, OAM differs from works \cite{chen2017sca,xu2015show,fu2017look} in that it is performed by the random walk graph propagation, which is in a parameter-free manner and can be directly employed without training. And different from \cite{chen2017sca,hu2017squeeze} where the channel-attention modules are to extract the most informative channel-signals, our parallel CAMs are constrained by the adversary task such that they are different with each other and capable of extracting rich and complementary attributes knowledge. In summary, CAMs can work in conjunction with OAMs, and the generalization and discrimination of the holistic metric are explicitly and naturally ensured.
\vspace{-1em}
\section{Experiments}\label{experiments}
\vspace{-0.5em}
\textbf{Notation}: For clarity, the conventional unified metric learning(i.e. directly train $d$-dim features with Eq.\ref{eq6}) is denoted by U$d$, e.g. U512, and we set it as our baseline. The proposed method is denoted by DeML(I,J), where there are $I$ OA root-learners and $I\times{J}$ CA sub-learners in total, d is set to 512 by default and each OA(CA) learner is $\frac{512}{I}$($\frac{512}{IJ}$)-dim, if not divisible, operation \emph{floor} is performed.

\textbf{Implementation details}: Following the codes released by \cite{chen2019energy,oh2016deep}, we choose the pretrained \emph{GooglenetV1} \cite{Szegedy2014Going} as our bedrock network and adopt the same data preprocess method so as to make fair comparisons with other works. The learners, CAMs and adversary networks are initialized with random weights. CAMs are placed after pool3 layer, OAM takes as inputs \emph{incep4e\_output} and \emph{incep5b\_output}.

\textbf{Training}: The optimizer is Adam \cite{kingma2014adam} with learning rate $\footnotesize{1e-5}$ and weight decay $\footnotesize{2e-4}$. The training iterations are $15k$(CUB), $25k$(CARS), $100k$(Stanford Online Products and In-shop) \emph{resp}. We set $\lambda_{0}=1,\lambda_{1}=0.014,\lambda_{2}=0.25$ and $10$ times learning rate for the learners.

\textbf{Evaluation and datasets}: For fair comparison, we follow \cite{oh2016deep} to evaluate the retrieval performance by Recall@K metric. For testing \emph{unseen} images, the outputs of $I\times{J}$ learners will be concatenated into the holistic feature after L2 normalization. Then we use the simple cosine distance rule for similarity measurement. The datasets include CUB\cite{Wah2011The}, CARS\cite{Krause20133D}, Stanford Online Products\cite{oh2016deep} and In-Shop\cite{liu2016deepfashion}. \textbf{There is no intersection of classes between training and testing sets.}
\subsection{Results}
\vspace{-0.5em}

\begin{figure}[t]
\vspace{-1.3em}
  \centering
  \includegraphics[width=0.8\linewidth]{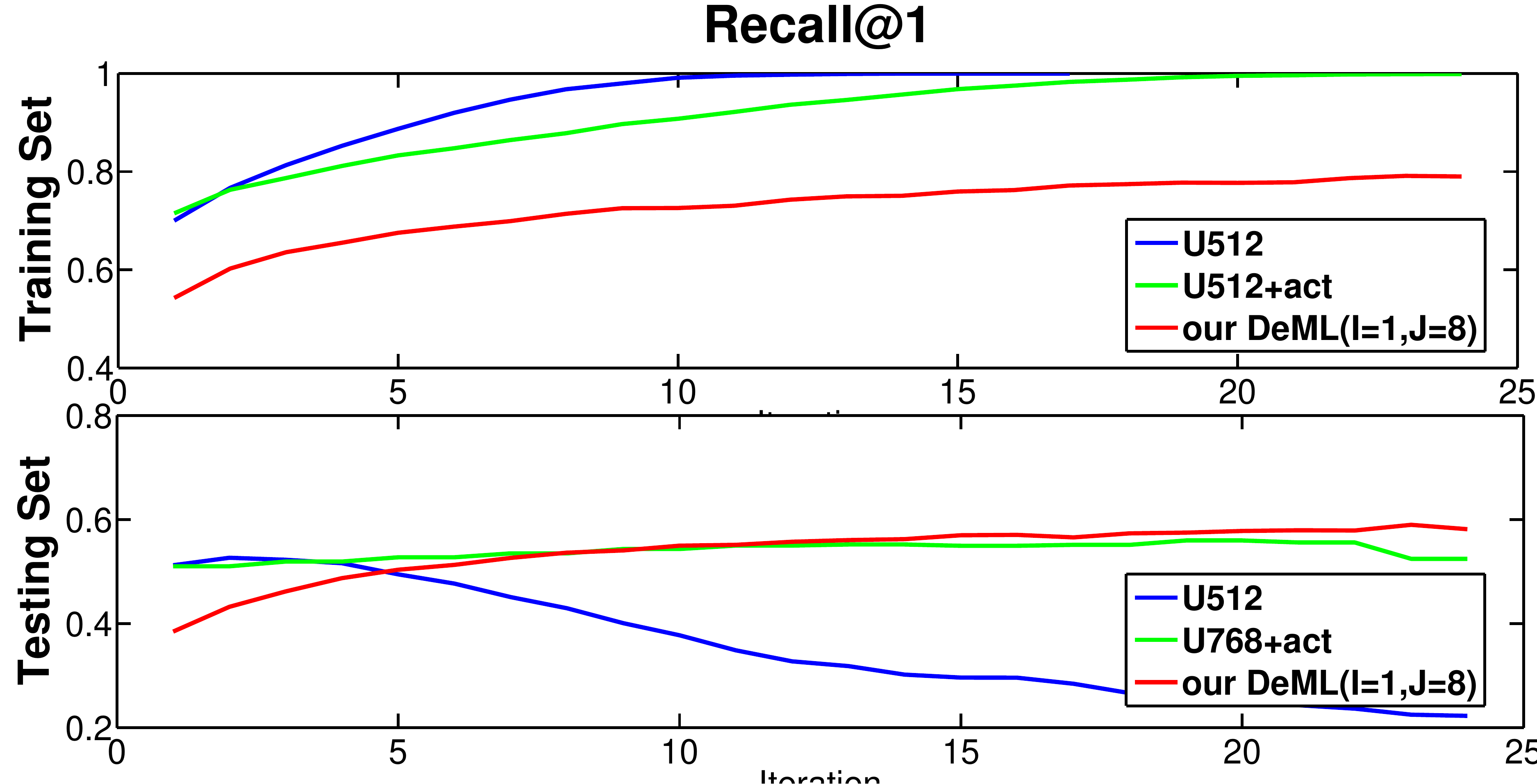}\\
    \vspace{-0.8em}\captionsetup{font={footnotesize}}
    \caption{Training (\emph{seen}) and testing (\emph{unseen}) curves on CUB.}\label{fig7}
    \vspace{-2em}
\end{figure}

\textbf{Effect of \emph{activation decay}}: From Tab.\ref{tab:addlabel5}, one can observe that with the help of our \emph{activation decay} $L_{act}$, the performances of unified metric learning under different feature sizes are consistently improved over both CUB
\begin{wraptable}{rt}{0.45\linewidth}
\hspace{-3em}
  \centering
  \resizebox{1.2\linewidth}{!}{
    \begin{tabular}{c|c|c}
    \hline
    Method & CUB (R@1) & CARS (R@1) \\
    \hline
    U64 & 50.0 & 65.4\\
    U64+$L_{act}$ & \textbf{50.9} $\uparrow$ & \textbf{67.1} $\uparrow$\\
    \hline
    U192  & 52.4  & 68.3 \\
    U192+$L_{act}$ & \textbf{54.9} $\uparrow$  & \textbf{74.6} $\uparrow$ \\
    \hline
    U384  & 53.0    & 69.1 \\
    U384+$L_{act}$ & \textbf{55.6} $\uparrow$  & \textbf{77.4} $\uparrow$ \\
    \hline
    U512  & 52.9  & 68.4 \\
    U512+$L_{act}$ & \textbf{56.2} $\uparrow$  & \textbf{77.6} $\uparrow$ \\
    \hline
    \end{tabular}%
    }\vspace{-0.5em}\captionsetup{font={small}}
      \caption{Effect of $L_{act}$.}
      \vspace{-1.5em}
  \label{tab:addlabel5}%
\end{wraptable}
and CARs datasets, e.g. (U512+$L_{act}$) \emph{vs.} U512, moreover, from Fig.\ref{fig7} one can observe that the training curve of U512 rises fast to a higher level, i.e. 1, but its testing curve first rises a bit and then drops continuously to 0.2, indicating that U512 suffers from serious over-fitting over the \emph{seen} categories, while the training curve of (U512+$L_{act}$) rises slower than U512 and its testing curve has a rising trend, implying that $L_{act}$ indeed serves as a regularization term and can improve the generalization ability on \emph{unseen} categories by reducing over-fitting. Additionally, from the results of (U64+$L_{act}$) to (U512+$L_{act}$), one can observe that the performances increase with dimension $d$, but when $d$ is large enough (e.g. from 384 to 512), the performance gain is small ($\footnotesize{55.6\%~vs.~56.2\%}$ on CUB, $\footnotesize{77.4\%~vs.~77.6\%}$ on CARS), and the similar phenomenon can be observed from U64 to U512 as well, indicating that the neural redundancy indeed exists and limits the further improvement, and highlighting the necessity of our decoupling idea laterally.

\textbf{Attention visualization}: In order to intuitively understand our attention modules, we provide clear visualizations of the corresponding attention proposals for qualitative analysis, i.e. the attended regions at multiple scales by the proposed OAMs and the diverse attribute proposals by CAMs, Fig.\ref{fig4}. Firstly, for object-attention regions, one can observe that these localized regions at second or third scales are discriminative to the corresponding categories, and are easier to be recognized than the first scale. Secondly, for channel-attention proposals, it can be observed that different CAMs process the same input $U^{l}$ into diverse proposals, which contain different attributes combinations, by different weighting strategies. Thus forcing the following CA learners to concentrate on different knowledge, the partial learning behavior can be avoided. These above results are consistent with human perception that it would be helpful to look closer for better decisions and to learn rich knowledge for handling \emph{unseen} categories.

\begin{table}[t]
\centering
\resizebox{0.75\linewidth}{!}{
    \begin{tabular}{c|c|c|c|c}
    \hline
          & \multicolumn{2}{c|}{CUB (R@1)} & \multicolumn{2}{c}{CARS (R@1)} \\
    \hline
    DeML & H     & O (1st/2nd/3rd) & H     & O (1st/2nd/3rd) \\
    \hline
    (I=1,J=1) & 56.1  & -     & 77.9  & - \\
    \hline
    (I=1,J=4) & 56.6  & -     & 81.1  & - \\
    (I=1,J=8) & 59.0  & -     & 82.5  & - \\
    \hline
    (I=2,J=1) & 60.9  & 54~/~58.8 & 83.4  & 76.9~/~79.2 \\
    (I=2,J=4) & 61.7  & 54.9~/~59.7 & 84.8  & 77.7~/~79.9 \\
    \hline
    (I=3,J=1) & 64.9  & 54~/~58~/~60.1 & 85.6  & 75.4~/~77.5~/~78.9 \\
    (I=3,J=3) & \textbf{65.4}  & 54.1~/~58.4~/~60.6 & \textbf{86.3}  & 75.3~/~78.9~/~80.4 \\
    \hline
    \end{tabular}%
    }%
\vspace{-0.6em}\captionsetup{font={scriptsize}}
      \caption{Effect of attention modules. H and O indicate the retrieval results of features coming from the \emph{holistic learner} and \emph{OA root-learners}, \emph{resp}.}\label{tab:addlabel6}
\end{table}
\begin{table*}[t]
\vspace{-2em}
  \centering
\resizebox{0.95\textwidth}{!}{
    \begin{tabular}{c|cccc|cccc|cccc|cccccc}
    \hline
    \multicolumn{1}{c|}{R@K(\%)} & \multicolumn{4}{c|}{CUB-200}      & \multicolumn{4}{c|}{CARS-196} & \multicolumn{4}{c}{Stanford Online Products}& \multicolumn{6}{c}{In-Shop}\\
    \hline
    Method   & 1     & 2     & 4     & 8     & 1     & 2     & 4     & 8 & 1     & 10    & 100   & 1000  & 1     & 10    & 20    & 30    & 40    & 50 \\
    \hline
    Lifted \cite{oh2016deep}   & 47.2  & 58.9  & 70.2  & 80.2  & 49.0    & 60.3  & 72.1  & 81.5 & 62.1  & 79.8  & 91.3  & 97.4 & -   & -    & -    & -    & -    & -\\
    N-pair \cite{Sohn2016npair}   & 51.0    & 63.3  & 74.3  & 83.2  & 71.1  & 79.7  & 86.5  & 91.6 & 67.7  & 83.8& 93.0& 97.8& -   & -    & -    & -    & -    & - \\
    Angular \cite{wang2017deep}  & 53.6  & 65.0   & 75.3  & 83.7  & 71.3  & 80.7  & 87.0    & 91.8 & 70.9  & 85.0    & 93.5  & 98.0& -   & -    & -    & -    & -    & -\\
    Proxy NCA \cite{Movshovitz-Attias_2017_ICCV}  & 49.2  & 61.9  & 67.9  & 72.4  & 73.2  & 82.4  & 86.4  & 88.7& 73.7  & -     & -     & - & -   & -    & -    & -    & -    & - \\
    ALMN \cite{chen2018almn} & 52.4 & 64.8 & 75.4 & 84.3 & 71.6 & 81.3 & 88.2 & 93.4 & 69.9 & 84.8 & 92.8 & - & -   & -    & -    & -    & -    & -\\
    ECAML \cite{chen2019energy} & 55.7 & 66.5 & 76.7 & 85.1 & 84.5 & 90.4 & 93.8 & 96.6 & 71.3 & 85.6 & 93.6 & 98.0 & 83.8   & 95.1    & 96.6    & 97.3    & 97.7    & 98.0\\
    HDC \cite{Yuan_2017_ICCV} $\circ$     & 53.6  & 65.7  & 77.0    & 85.6  & 73.7  & 83.2  & 89.5  & 93.8& 69.5  & 84.4  & 92.8  & 97.7 & 62.1  & 84.9  & 89.0    & 91.2  & 92.3  & 93.1 \\
    BIER \cite{Opitz_2017_ICCV} $\circ$   & 55.3  & 67.2  & 76.9  & 85.1  & 78.0    & 85.8  & 91.1  & 95.1& 72.7  & 86.5  & 94.0    & 98.0 & 76.9  & 92.8  & 95.2  & 96.2  & 96.7  & 97.1 \\
    ABE \cite{Kim_2018_ECCV}\textsuperscript{\ref{ABE}} $\circ$ & 58.6 & 69.9 & 79.4 & 87.1 & 82.7 & 88.8 & 93.1 & 96.0 & 74.7 & 87.9 & 94.2 & 98.0& 86.2 & 96.1 & 97.5 & 97.9 & 98.2 & 98.5\\
    A-BIER \cite{opitz2018deep} $\circ$   & 57.5  & 68.7  & 78.3  & 86.0  & 82.0    & 89.0  & 93.2  & 96.1& 74.2 & 86.9  & 94.0    & 97.8 & 83.1  & 95.1  & 96.9  & 97.5  & 97.8  & 98.0\\
    \hline
    baseline(U512) & 52.9  & 65.0  & 75.4 &  83.6 & 68.4  & 78.7  & 86.0  & 91.2 & 68.7  &  84.0 & 92.7 & 97.6 &  81.8 & 94.1 & 95.7 & 96.5 & 97 & 97.3 \\
    \bf{\emph{DeML(I=3,J=3)}}  & \textcolor[rgb]{1.000, 0.000, 0.000}{\bf{\emph{65.4}}}  & \textcolor[rgb]{1.000, 0.000, 0.000}{\bf{\emph{75.3}}}  & \textcolor[rgb]{1.000, 0.000, 0.000}{\bf{\emph{83.7}}}  & \textcolor[rgb]{1.000, 0.000, 0.000}{\bf{\emph{89.5}}}  & \textcolor[rgb]{1.000, 0.000, 0.000}{\bf{\emph{86.3}}}  & \textcolor[rgb]{1.000, 0.000, 0.000}{\bf{\emph{91.2}}}    & \textcolor[rgb]{1.000, 0.000, 0.000}{\bf{\emph{94.3}}}  & \textcolor[rgb]{1.000, 0.000, 0.000}{\bf{\emph{97.0}}} & \textcolor[rgb]{1.000, 0.000, 0.000}{\textbf{\emph{76.1}}}$^{*}$  & \textcolor[rgb]{1.000, 0.000, 0.000}{\textbf{\emph{88.4}}}$^{*}$  & \textcolor[rgb]{1.000, 0.000, 0.000}{\textbf{\emph{94.9}}}$^{*}$  & \textcolor[rgb]{1.000, 0.000, 0.000}{\textbf{\emph{98.1}}}$^{*}$& \textcolor[rgb]{1.000, 0.000, 0.000}{\textbf{\emph{88.2}}}  & \textcolor[rgb]{1.000, 0.000, 0.000}{\textbf{\emph{97.0}}}    & \textcolor[rgb]{1.000, 0.000, 0.000}{\textbf{\emph{98.0}}}    & \textcolor[rgb]{1.000, 0.000, 0.000}{\textbf{\emph{98.3}}}  & \textcolor[rgb]{1.000, 0.000, 0.000}{\textbf{\emph{98.6}}}  & \textcolor[rgb]{1.000, 0.000, 0.000}{\textbf{\emph{98.8}}}\\
    \hline
    \end{tabular}%
}\captionsetup{font={footnotesize}}\vspace{-1em}
      \caption{Retrieval results on CUB \cite{Wah2011The}, CARS \cite{Krause20133D} and Stanford Online Products \cite{oh2016deep}. Here, $\circ$ refers to other ensemble methods, the superscript $^{*}$ indicates the results are from model DeML(I=2,J=4) not DeML(I=3,J=3).}
      \vspace{-1em}
  \label{tab:addlabel1}%
\end{table*}%

\textbf{Effect of attention modules}: We also conduct quantitative comparisons on attention modules in Tab.\ref{tab:addlabel6}. By default, dimension $d$ is set to 512. The model DeML(I=1,J=1) is very similar to model (U512+$L_{act}$) with only a small difference of an extra single CAM, and from Tab.\ref{tab:addlabel5} and Tab.\ref{tab:addlabel6}, one can observe that their performances are almost the same($56.2\%$ \emph{vs}. $56.1\%$ on CUB, $77.6\%$ \emph{vs.} $77.9\%$ on CARS), implying that capturing the single attentional knowledge via attention module might not be helpful for ZSIR. Moreover, in Tab.\ref{tab:addlabel6} one can observe that, by fixing J=1, the holistic performance (H) will significantly increase with I ($56.1\%\rightarrow60.9\%\rightarrow64.9\%$ on CUB, $77.9\%\rightarrow83.4\%\rightarrow85.6\%$ on CARS), and the OA learner at finer scale will be more discriminative than that at coarse scale, e.g. in DeML(I=3,J=1) on CUB, the third OA learner achieves $60.1\%$, the second gets $58\%$ and the first gets $54\%$, indicating that the more discernible the input is, the more discriminative embedding is. Furthermore, when fixing I to a certain value, decoupling each OA root-learner into multiple CA sub-learners will consistently improve the performances over both CUB and CARS, e.g. when I=1, the result (H) increases with J from $56.1\%$ to $59.0\%$ ($77.9\%$ to $82.5\%$) on CUB (CARS), and the similar improvements can be also observed when I=2 or 3, revealing that explicitly encouraging the diversity and capturing rich knowledge indeed improve the generalization of metric in ZSIR. This conclusion can also be demonstrated by Fig.\ref{fig7}, i.e. the training curve of DeML(I=1,J=8) is much lower than that of U512 or (U512+$L_{act}$), showing that the learned knowledge are not specially prepared for the training seen categories and the partial learning behavior is mitigated, while its testing curve outperforms both U512 and (U512+$L_{act}$), implying that learning diverse knowledge will further improve the generalization ability by reducing over-fitting. Worthy of mention is that in theory bigger J will be more helpful(e.g. J=8), however due to the limitation of GPU memory, during training when I=3, J is limited to 3.
In summary, the appropriate combination of (I,J) will significantly improve both the discrimination and generalization of the deep metric in an interpretable attention-based decoupling manner.


\textbf{Effect of adversary loss}: Unlike OAMs that can automatically extract the object-attention regions via random-walk graph propagation, the CAMs should be extra constrained so as to explicitly encourage the learning of complementary attributes knowledge instead of the partial one. From Tab.\ref{tab:addlabel7}, one can observe that without the constraint of $L_{adv}$, the performances of DeML(I=1,J=8) are drastically decreased over both CUB and CARS (i.e. from $59.0\%$ to $52.0\%$ on CUB and $82.5\%$ to $68.2\%$ on CARS), and the similarities between different CA learners are obviously increased from 0 to nearly 1, verifying that our adversary loss $L_{adv}$ indeed encourages the discrepancies between CA learners (and CAMs) and reduces the neural redundancy, as a result, the diverse and rich information are captured such that the generalization ability is enhanced. Moreover, one can observe that even if there are 8 64-dim learners in (DeML(I=1,J=8) without $L_{adv}$), the performances are close to (U64+$L_{act}$), since these 8 learners are supervised independently and thus are all likely to focus on the same attributes knowledge that are the easiest one to reduce the training risk, and then have little differences with each other('sim' is close to 1), implying the redundant learners will not contribute much and the partial learning behavior will turn up easily without the diversity constraint.
\begin{table}[t]
  \centering
  \vspace{-1em}
  \resizebox{0.95\linewidth}{!}{
    \begin{tabular}{c|cccc|c}
    \hline
    CUB (R@K)   & 1     & 2     & 4     & 8 & sim\\
    \hline
    DeML(I=1,J=8) & \textbf{59.0}  & \textbf{70.2}  & \textbf{79.7}  & \textbf{87.4} & -0.08 \scriptsize${\pm0.01}$\\
    DeML(I=1,J=8) without $L_{adv}$ & 52.0 $\downarrow$    & 63.9 $\downarrow$  & 74.0 $\downarrow$    & 82.3 $\downarrow$ & 0.92 \scriptsize{${\pm0.02}$} $\uparrow$\\
    U64+$L_{act}$ & 50.9  & 62.8  & 74.0    & 82.6 & -\\
    \hline
    CARS (R@K)   & 1     & 2     & 4     & 8 & sim\\
    \hline
    DeML(I=1,J=8) & \textbf{82.5}  & \textbf{89.5}  & \textbf{93.5}  & \textbf{96.1} & -0.09 \scriptsize${\pm0.01}$\\
    DeML(I=1,J=8) without $L_{adv}$ & 68.2 $\downarrow$  & 78.3 $\downarrow$  & 85.4 $\downarrow$  & 90.9 $\downarrow$ &0.95 \scriptsize{${\pm0.01}$} $\uparrow$\\
    U64+$L_{act}$ & 67.1  & 77.6  & 85.0    & 90.6 &-\\
    \hline
    \end{tabular}%
}\vspace{-0.5em}
    \captionsetup{font={scriptsize}}
      \caption{Effect of adversary loss. 'sim' indicates the mean cosine similarity between CA learners.}
\vspace{-2em}
  \label{tab:addlabel7}%
\end{table}%

\textbf{Comparison with State-of-the-art}: In order to highlight the significance of our decoupling idea for \emph{zero-shot} retrieval, we compare DeML with some remarkable embedding works
\footnote{For fair comparison, we report the reproducing results of \cite{Kim_2018_ECCV} when using the same training settings as us, e.g. batchsize and loss function.\label{ABE}}, over the popular used datasets CUB, CARS, Stanford Online Products and In-Shop Clothes. The comparison results are listed in Tab.\ref{tab:addlabel1}. From these tables, one can observe that our baseline(U512) can only achieve the general performances, and the performances of traditional ideas of loss-designing and samples-mining are almost on par with each other, while, by explicitly intensifying both the discrimination and diversity within the deep metric, our DeML can significantly improve the performances over baseline model and outperforms other state-of-the-art methods by a noteworthy margin, demonstrating the necessity of our explicit enhancement for discrimination and generalization via the decoupling idea. Moreover, different from the listed ensemble methods \cite{Yuan_2017_ICCV,Opitz_2017_ICCV,opitz2018deep,Kim_2018_ECCV}, DeML has clear objects of jointly mitigating the aforementioned issues that are vital to ZSIR, and thus can easily surpass them. Worthy of mention is that, we find DeML(I=2,J=4) is enough for Stanford Online Products dataset, since as in Fig.\ref{fig4} the second scale has been capable of localizing the discriminative regions.

\vspace{-0.8em}
\section{Conclusion}
\vspace{-0.8em}
In this paper, we propose an interpretable DeML, a hybrid-attention based decoupling method, for \emph{zero-shot image retrieval} task by explicitly intensifying the discrimination and generalization within the learned metric with the proposed OAMs and CAMs. Extensive experiments validate the necessity and effectiveness of our decoupling idea. 

\small{\textbf{Acknowledgments}: This work was partially supported by the National Natural Science Foundation of China under Grant Nos. 61871052, 61573068, 61471048, and 61375031, and by the Beijing Nova Program under Grant No. Z161100004916088, DiDi GAIA Research Collaboration Initiative and BUPT Excellent Ph.D. Students Foundation CX2019307.}
{\small
\bibliographystyle{ieee}
\bibliography{egbib}
}

\end{document}